# Counterfactual Tool Ranking under Utility, Cost, and Privilege Constraints

## Realized-Return Controls, Public Tool-Calling Data, and Disagreement-Support Evaluation

**Jiapeng Li** (Microsoft)

## Abstract

Counterfactual tool evaluation must distinguish authority, historical support, and what a comparison actually estimates. We study these distinctions with eleven executable enterprise-inspired tools, exact-propensity logs, and real local Model Context Protocol transport. An initial 45-run synthetic study is retained, then challenged by 30 realized-return control runs and 15 experiments on 1,930 independently released Berkeley Function Calling Leaderboard (BFCL) tasks. Full-return direct regression reverses an initially favorable doubly robust (DR) evaluation result in the linear setting: mean absolute errors are 0.0139 for direct regression and 0.0272 for DR. Under a shifted environment, DR retains an advantage, with errors 0.0227 versus 0.0948. On function-name-group-disjoint BFCL-derived splits, direct and DR selectors obtain balanced accuracies of 81.85% and 79.83%. Two pinned local Qwen2.5 models are evaluated on the same 200 held-out tasks, exposing a strong failure to abstain under the fixed prompt. We further characterize policy differences under missing support: unsupported actions shared by two policies cancel, allowing point identification of an incremental change when neither absolute value is identifiable. A disagreement-preserving fallback achieves this property in all five support-gap runs, but conservative sampling bounds do not certify deployment improvement. The contribution is a falsifiable evaluation method and independent public evidence, not a new DR estimator, official BFCL leaderboard score, or production-agent safety claim.

## 1. Introduction

An enterprise agent may have access to many tools but cannot legitimately use every tool on every resource. Reading one document, exporting a collection, closing a ticket, and applying a customer discount can all be semantically relevant while differing sharply in scope, cost, and consequences. Ranking tools by textual relevance alone does not decide whether a cheap result is stale, whether a write requires a current approval, or whether a broader operation unnecessarily accesses other records.

There is a useful connection to advertising systems. A request and its observed state form a context; candidate tool calls form actions; the deployed selection policy generates partial feedback; and a new policy must be evaluated without pretending that unexecuted alternatives have observed labels. The connection is not an equivalence. A click has no direct counterpart to a revoked permission, and changing an agent's first action can alter every subsequent state. A single-decision estimator cannot simply be applied to a complete multi-step trajectory by copying the terminal success label onto each step. This framing adapts the established contextual-bandit and logged-feedback perspective [1,2] to complete agent tool calls; we do not claim the underlying reduction as new.

We therefore deliberately study a narrower, identifiable problem: selection among fixed complete tool-call candidates at one decision point. We control the candidate generator, record exact sampling probabilities after permission filtering, and evaluate fixed policies in independently reset environments. No language model is used to infer an authorization decision or to judge task success. The initial study isolates learning from language parsing; the new public-data study separately evaluates natural-language function selection with local LLMs and label-free lexical features. It still does not claim to solve long-horizon credit assignment.

The empirical questions are:

1. Does counterfactual learning improve utility over strong direct-prediction and hand-written baselines under identical action and permission boundaries?
2. How accurately do direct, IPS, self-normalized IPS, and DR estimators predict the actual values of frozen policies?
3. What happens when exploratory data are scarce, actions have zero historical probability, observations are stale, or the environment changes?
4. What utility and coverage are lost when a policy requires local evidence before acting?
5. Can a policy difference remain identifiable when its two absolute values are not, and does the distinction help preserve useful baseline behavior?

The version-2 results reject blanket superiority claims about either DR ranking or DR evaluation. They also correct a weakness in our initial comparison: nominal-cost direct prediction is not a substitute for regression of the full realized return. Sections 3--7 retain the first study and its frozen artifacts; Section 8 adds the new controls, public dataset, actual local LLM runs, and disagreement-support analysis. The old data are not silently overwritten.

## 2. Related Work and Scope

Doubly robust policy evaluation and learning combine a reward model with propensity weighting [1]. Counterfactual risk minimization formalizes learning from logged bandit feedback and highlights variance-sensitive objectives [2]. Support deficiency is a known limitation of off-policy bandits, with established responses including action restriction, policy restriction, and extrapolation under additional assumptions [3]. Safe policy improvement with baseline bootstrapping provides a different, theoretically grounded response to uncertain regions [4]. Our evidence threshold and ensemble disagreement are not an implementation of that theorem and do not inherit its guarantees.

Progent constrains tool names and arguments with deterministic policy checks, and its current revision distinguishes policy narrowing from expansions that require approval [5]. MiniScope addresses least-privilege authorization and permission hierarchies [6]. These systems motivate treating authorization as an external execution constraint, not as a utility penalty that a model may trade away. We do not claim a new access-control mechanism or reproduce their attack benchmarks.

Tau-bench evaluates tool-agent-user interaction through task and database-state outcomes [7]. AgentAbstain studies paired should-act and should-abstain tasks [8]. Our small generated tasks borrow the principle of executable outcome checking, not their datasets or reported scores. We have not run either external benchmark. Sequential DR estimators exist for reinforcement learning [9]; their assumptions and trajectory weighting are outside the present single-decision study. Open Bandit Dataset and Pipeline exemplify reproducible comparison of off-policy estimators using multiple logged policies [10].

Thus the combination of an ACL, DR estimation, and abstention is not presented as algorithmic novelty. The artifact contributes an explicit contract connecting these components, failure-revealing conditions, real MCP transport parity, and results that can be regenerated without paid APIs or private data.

The extension uses BFCL's independently collected live function-call queries [11], not only our own generator, and two small Qwen2.5 models [12]. Our disagreement-support proposition follows from a linear policy contrast and bounded missing outcomes. It is closely related to deficient-support and baseline-restriction ideas [3,4]; we make no unsupported claim of priority over general partial-identification theory. Its specific contribution here is the contrast-level support contract, executable bounds, and a baseline-preserving comparison that can be falsified on public tool-call data. This is stronger than simply naming the combination of DR and ACL a new algorithm, but does not by itself settle the contribution's ultimate novelty or impact.

# 3. Problem Formulation

## 3.1 Complete actions and feasible sets

Let $x$ contain the request's structured pre-action observations. Let $P$ be the trusted authorization snapshot and $C(x)$ the fixed candidate set. An action $a$ contains a tool identifier, an explicit list of resource identifiers, and, when applicable, an integer discount amount. The authorized candidate set is

$$A(x, P) = \{a \in C(x) : \mathrm{Auth}(P, a) = 1\} \cup \{\perp\},$$

where $\perp$ denotes abstention. The ranking policy satisfies $\pi(a \mid x, P, C) = 0$ outside this set. For compactness, subsequent expressions write the complete observed decision context, including the candidate set and permission snapshot, as $z$.

The candidate generator is fixed for every compared policy. It supplies exact resource templates rather than learning free-form arguments. A separate unauthorized cross-tenant candidate is included before masking. The benchmark therefore tests filtering complete calls, not merely approving a tool name.

## 3.2 Utility and non-negotiable authorization

For an executed action, the measured utility is

$$r = s - \lambda_c c - \lambda_l \frac{\ell}{100} - \lambda_u u - 0.05e,$$

where $s$ is verified task success, $c$ is simulated service cost, $\ell$ is simulated latency in milliseconds, $u$ indicates an unsafe business outcome, and $e$ counts supplied resource arguments outside the requested resource set. The default weights are $\lambda_c = 1$, $\lambda_l = 0$, and $\lambda_u = 2$. Abstention yields $s = r = 0$. Denied actions perform no mutation and incur no service fee in this benchmark. An injected service failure incurs the nominal fee and twice the nominal simulated latency, but occurs before mutation.

Authorization and $u$ are different axes. A principal may be authorized to write a ticket while closing it without its current approval is still an unsafe business outcome. The ACL is a hard gate; business risk remains part of the utility and evaluation. We do not allow a high predicted reward to bypass an ACL denial. The extra-resource term is a simple declared-argument exposure proxy, not a measure of data sensitivity or a proof of least privilege.

## 3.3 Partial feedback and support

Each log entry contains $(z_i, a_i, \mu_i, r_i)$, together with the complete post-mask probability vector and reproducible environment state. Here $\mu_i = \mu(a_i \mid z_i)$ is the actual action-sampling probability, not a language-model confidence or a fitted propensity. The learning interface only receives the selected action's result.

Identifying a target value requires overlap:

$$\pi(a \mid z) > 0 \implies \mu(a \mid z) > 0.$$

Permission does not imply overlap. A tool can be currently authorized but never explored by the logging policy. In that case our evaluator reports the full target value as not identified. It does not drop those contexts, silently renormalize, or substitute a reward-model prediction as identified evidence. Estimating a value there would require explicit additional assumptions.

# 4. System and Threat Model

## 4.1 Authorization path

The policy contains a principal, groups, scopes, allowed resource prefixes, explicit denied prefixes, and subject-specific scope/resource grants. Prefix inheritance respects path segment boundaries. A matching user or group deny takes precedence; an allow must match the same scope and resource. Grants are not flattened into independent scope and resource sets, which would introduce an unintended Cartesian product of authority. Unknown tools and malformed resource identifiers fail closed. Discount arguments must be genuine integers within 1 through 100; booleans and strings are rejected at the MCP boundary.

Every candidate is filtered before ranking. Immediately before execution, the same complete action is checked against the latest trusted execution policy. The shifted condition can revoke authority between the ranking snapshot and execution. Group membership and policies are supplied by the trusted fixture, not by the tool caller.

The invariant is conditional: if all reads and writes go through this mediator, the authoritative policy is correct, and no alternate execution path exists, then an action denied by the execution policy cannot perform a sandbox read or mutation. Tests inspect both returned outcomes and database state. This is not a proof of an external tool's implementation, identity provider, operating-system isolation, or concurrent production authorization transaction. The benchmark holds the trusted policy immutable for a call; it tests snapshot-to-execution revocation, not arbitrary mid-write races.

## 4.2 Real protocol boundary, synthetic services

Two backends share the same domain implementation. The local backend invokes the sandbox directly. The MCP backend uses the official Python SDK 2.2.0 to launch an independent stdio server, discover its tool schemas, and make typed tool calls. The server loads tasks and policies from a trusted local manifest at startup. Callers submit task IDs and action arguments, not arbitrary policy objects or executable commands. No service listens on a network interface.

Each call creates fresh in-memory SQLite state. This makes policy comparisons independent and replayable. It does not model a persistent MCP user session or cross-call side effects. Test fixtures and hidden failure realizations are available to the trusted execution harness, but the ranker's feature allowlist excludes them. In particular, true approval state, actual failed tools, execution-time revocation, task IDs, and database content are not model features.

## 4.3 Logging and reproducibility

The logger is epsilon-greedy over authorized actions: it chooses the cheapest executable action with probability $1 - \epsilon$ and explores uniformly over the eligible authorized set, including abstain, with probability $\epsilon$. If no executable action exists, it chooses abstain. In the deficient-support condition, designated tools remain candidates but receive logging probability zero. Only one action is executed per collection decision.

Schema-v2 JSONL records include the full candidate/authorization snapshot, resource and numeric arguments, probability vector, selected probability, policy snapshots, tool-catalog hash, selected result, and task seed metadata. Readers reject mismatched candidates, changed tool catalogs, nonfinite outcome metrics, malformed flags, and probabilities inconsistent with the declared logger. Replays compare semantic outcomes while excluding measured runtime. Hashes support provenance and change detection; they are not cryptographic attestations that an untrusted logger told the truth.

# 5. Learning and Evaluation Methods

## 5.1 Direct and counterfactual score models

All learned methods receive the same structured features: domain, requested field and count, freshness requirement, observed approval and discount limit, observed service health, cache-age category, tool identity, argument count, discount amount, and available scopes. No text embedding or hidden database label is used. Direct learning predicts success and unsafe-outcome probabilities and subtracts nominal cost, latency, and argument-exposure penalties. Values of probability predictions are clipped to $[0, 1]$.

For DR learning, three-fold cross-fitting produces an out-of-fold outcome estimate $\hat{q}_{-k(i)}(z_i, a)$. Every supported candidate receives the target

$$\tilde{r}_i^{\mathrm{DR}}(a) = \hat{q}_{-k(i)}(z_i, a) + \frac{\mathbf{1}\{a_i = a\}}{\mu(a_i \mid z_i)} \left[r_i - \hat{q}_{-k(i)}(z_i, a_i)\right] .$$

A separate model regresses these targets and selects the highest-scoring legal action. The IPS learner uses $\tilde{r}_i^{\mathrm{IPS}}(a) = \mathbf{1}\{a_i = a\} r_i / \mu(a_i \mid z_i)$. Unsupported candidate rows are excluded. Each expanded row has weight $1/|A(z_i)|$; thus each full-support context has total weight one, while a deficient-support context retains only its supported fraction of that weight. We do not claim that this regression reduction is a new policy optimizer.

Tree experiments use Extra Trees with 48 estimators, depth at most 12, and a minimum of three observations per leaf. The linear ablation uses Ridge regression with regularization 10. Model-family settings are shared across learned policies, although direct learning has two outcome heads and clipping, while IPS and DR each regress a scalar target. They are not literally identical function classes after all transformations.

## 5.2 Evidence-aware selective execution

The supported set used for deployment is a finite feature-stratum lookup over training logs, not a learned guarantee for arbitrary new contexts. The conservative variants additionally require at least five selected observations for the context/action stratum. They rank by predicted value minus an ensemble disagreement penalty, and abstain if this score does not exceed a threshold.

Direct and DR conservative policies use the same evidence floor and independent calibration over disagreement weights $\{0, 0.5, 1\}$ and thresholds $\{0, 0.5, 0.8, 0.9, 1\}$. The selected pair maximizes the lower endpoint of a task-bootstrap DR interval on a separate calibration split. A fully unsupported calibration set falls back to abstention. Linear models have a single estimator, so their ensemble-disagreement term is zero.

This procedure is intentionally described as a heuristic. Tree disagreement is not a calibrated conditional confidence bound. Reusing a calibration set to select among multiple intervals does not give a uniform safe-improvement guarantee. The evidence threshold can reject useful actions simply because structured contexts are sparse. Comparing conservative direct and conservative DR prevents such rejection from being attributed uniquely to counterfactual learning.

### 5.3 Off-policy estimators

For a frozen deterministic target $\pi$, let $w_i = \mathbf{1}\{a_i = \pi(z_i)\}/\mu(a_i \mid z_i)$. We compare direct prediction, IPS, self-normalized IPS, and DR:

$$\widehat{V}_{\mathrm{DM}} = \frac{1}{n}\sum_i \hat{q}(z_i, \pi(z_i)), \qquad \widehat{V}_{\mathrm{IPS}} = \frac{1}{n}\sum_i w_i r_i,$$

$$\widehat{V}_{\mathrm{SNIPS}} = \frac{\sum_i w_i r_i}{\sum_i w_i},$$

$$\widehat{V}_{\mathrm{DR}} = \frac{1}{n}\sum_i \left[\hat{q}(z_i, \pi(z_i)) + w_i(r_i - \hat{q}(z_i, a_i))\right].$$

The OPE outcome model is trained only on the training split; target policies and thresholds are fixed before test feedback is evaluated. No propensity clipping is applied. We report effective sample size $\mathrm{ESS} = (\sum_i w_i)^2 / \sum_i w_i^2$ and warn about low ESS or no matches. A zero-match DR estimate is entirely model-based for that sample, even when population support is positive.

Under randomized logging, correct probabilities, overlap, and an independent outcome model, the conditional expectation of the residual term is $q(z, \pi(z)) - \hat{q}(z, \pi(z))$. It cancels the direct-model error. This is the standard DR identity [1], not a new theorem. Finite-sample variance, regression approximation, and the nonlinear maximization step explain why a good value estimator need not produce the strongest learned ranking policy.

## 6. Experimental Design

### 6.1 Executable workloads

Document tasks request titles or content for one or three records, sometimes requiring current data. Tools provide title search, cached reads, one-record live reads, batch reads, and broader exports. Success is checked against actual returned fields and freshness requirements. A one-record tool cannot satisfy a three-record task.

Ticket tasks close one or three records subject to current approval. The checked tool validates all approvals before a transactional write; cheap and administrative variants can produce an unsafe mutation. Customer-discount tasks additionally require an integer requested percentage and per-record discount limits. Their checked variant is likewise atomic. Success uses the database's terminal values and rejects wrong-resource or unapproved changes.

Tasks sample three domains uniformly, one or three target records uniformly, and a small finite set of numeric amounts and limits. The principal has a baseline scope set, optional broader scopes, and possible explicit denials. Nominal costs range from 0.01 to 0.16 simulated units; nominal latency ranges from 5 to 45 milliseconds. The all-abstain policy provides a zero-utility floor. The hindsight oracle executes every legal candidate in independent resets and selects the highest realized utility. It knows latent failures and is an upper bound, not a deployable competitor.

In noisy conditions, approval observations are flipped with probability 0.15, discount-limit observations may be stale, service-health reports are imperfect, cache freshness depends on observed age, and faults are sampled per tool before the task is logged. A shifted condition increases stale-observation probability to 0.30 and service degradation probability from 0.25 to 0.55, lowers cache freshness, and independently revokes execution authority on 12% of tasks. Failures are fixed potential outcomes per fixture, not newly sampled for each policy. This permits paired comparisons without exposing those outcomes during learning. Business feasibility excludes ACL/approval/limit violations but does not condition on realized service outages; our primary result is all-task success rather than the feasibility-normalized score.

### 6.2 Frozen matrix and baselines

The matrix comprises clean, noisy, small-data, low-exploration, missing-support, shifted, linear-model, cost-sensitive, and latency-sensitive conditions. Each uses seeds 7, 17, 23, 31, and 47. Defaults are 6,000 training decisions, 1,200 calibration decisions, 2,000 test tasks, and $\epsilon = 0.3$. Small-data runs use 600 training decisions; low-exploration runs use $\epsilon = 0.05$. Missing-support runs exclude the document batch tool and the checked ticket and discount tools from logging. Cost-sensitive runs use $\lambda_c = 3$; latency-sensitive runs use $\lambda_l = 0.5$.

Baselines include cheapest legal action, always selecting the complete batch/checked tool (`schema_match`), nominal hand-written rules, direct learning, IPS learning, DR learning, both conservative variants, abstain, and the oracle. The hand-written rules know nominal semantics and use observed preconditions, but do not know hidden faults or stale state. There is no LLM baseline; the `schema_match` label is not a claim about embedding retrieval or language-model performance.

All five seeds and every condition are retained. The matrix yields 387,000 selected-action log records and 90,000 held-out task-condition instances. These are not that many unique independent tasks: settings deliberately reuse seeds and some generated fixtures for paired comparisons. Splits are independently seeded within each run, but they draw from the same templates and do not establish unseen-template or unseen-tenant generalization. Source hashes and per-run summaries are published.

### 6.3 Metrics and statistical reporting

We report execution coverage, success over all tasks, success conditional on execution, cost per successful task including fees from failed attempts, simulated latency, unsafe-outcome rate, argument exposure, and mean utility. No successful execution is credited to abstention. Attempts denied at execution are reported separately. Authorization violations are not an exchangeable dimension of the utility Pareto frontier.

Per-run estimates use 200 task-level bootstrap resamples. Tables show mean and sample standard deviation across five seeds. These are descriptive results, not corrected significance tests or confidence certificates for rare safety events. OPE errors are computed for the same identified target-policy cases within each setting; abstain and the oracle are excluded from aggregate OPE comparisons. Utility values should be compared within a setting because some settings deliberately change the objective weights.

## 7. Initial Results: Frozen Version 1

### 7.1 Policy learning does not uniformly favor DR

Table 1 reports utility. Standard deviations across seeds appear in parentheses.

| Setting | Rules | Direct | IPS | DR | Cons. direct | Cons. DR |
|---|---|---|---|---|---|---|
| clean | 0.5923 (0.0085) | 0.5922 (0.0084) | 0.5280 (0.0315) | 0.5852 (0.0083) | 0.5286 (0.0155) | 0.5261 (0.0163) |
| cost sensitive | 0.4323 (0.0098) | 0.4243 (0.0087) | 0.3703 (0.0254) | 0.4199 (0.0114) | 0.2992 (0.0138) | 0.2897 (0.0155) |
| latency sensitive | 0.4421 (0.0102) | 0.4355 (0.0098) | 0.3869 (0.0167) | 0.4332 (0.0117) | 0.3042 (0.0149) | 0.2928 (0.0186) |
| linear | 0.4970 (0.0108) | 0.4662 (0.0122) | 0.4786 (0.0115) | 0.4778 (0.0106) | 0.3043 (0.0124) | 0.2972 (0.0192) |
| low exploration | 0.4970 (0.0108) | 0.4866 (0.0154) | 0.4371 (0.0210) | 0.4750 (0.0210) | 0.2519 (0.0470) | 0.2517 (0.0478) |
| missing support | 0.4970 (0.0108) | 0.3631 (0.0156) | 0.3026 (0.0202) | 0.3563 (0.0156) | 0.2590 (0.0130) | 0.2587 (0.0136) |
| noisy | 0.4970 (0.0108) | 0.5118 (0.0072) | 0.4765 (0.0255) | 0.5027 (0.0104) | 0.3409 (0.0156) | 0.3359 (0.0212) |
| shifted | 0.3509 (0.0108) | 0.3926 (0.0122) | 0.3663 (0.0047) | 0.3814 (0.0064) | 0.2492 (0.0077) | 0.2425 (0.0175) |

| Setting | Rules | Direct | IPS | DR | Cons. direct | Cons. DR |
|---|---|---|---|---|---|---|
| small data | 0.4970 (0.0108) | 0.4616 (0.0349) | 0.3817 (0.0532) | 0.4212 (0.0263) | 0.1165 (0.0140) | 0.1208 (0.0190) |

In the noisy tree setting, direct learning achieves 0.5118, compared with 0.5027 for DR and 0.4970 for rules. With small training data, direct and DR utilities are 0.4616 and 0.4212. The inverse-propensity correction is therefore not a free improvement in finite samples. Plain IPS regression is particularly variable and often produces unsafe business actions even in a clean environment.

The linear setting is an exception: DR reaches 0.4778 versus 0.4662 for direct learning, a mean paired difference of 0.0117 across seeds. However, IPS reaches 0.4786, slightly above DR. This supports a limited conclusion about the value of counterfactual targets under restricted model capacity, not a claim of unique DR superiority.

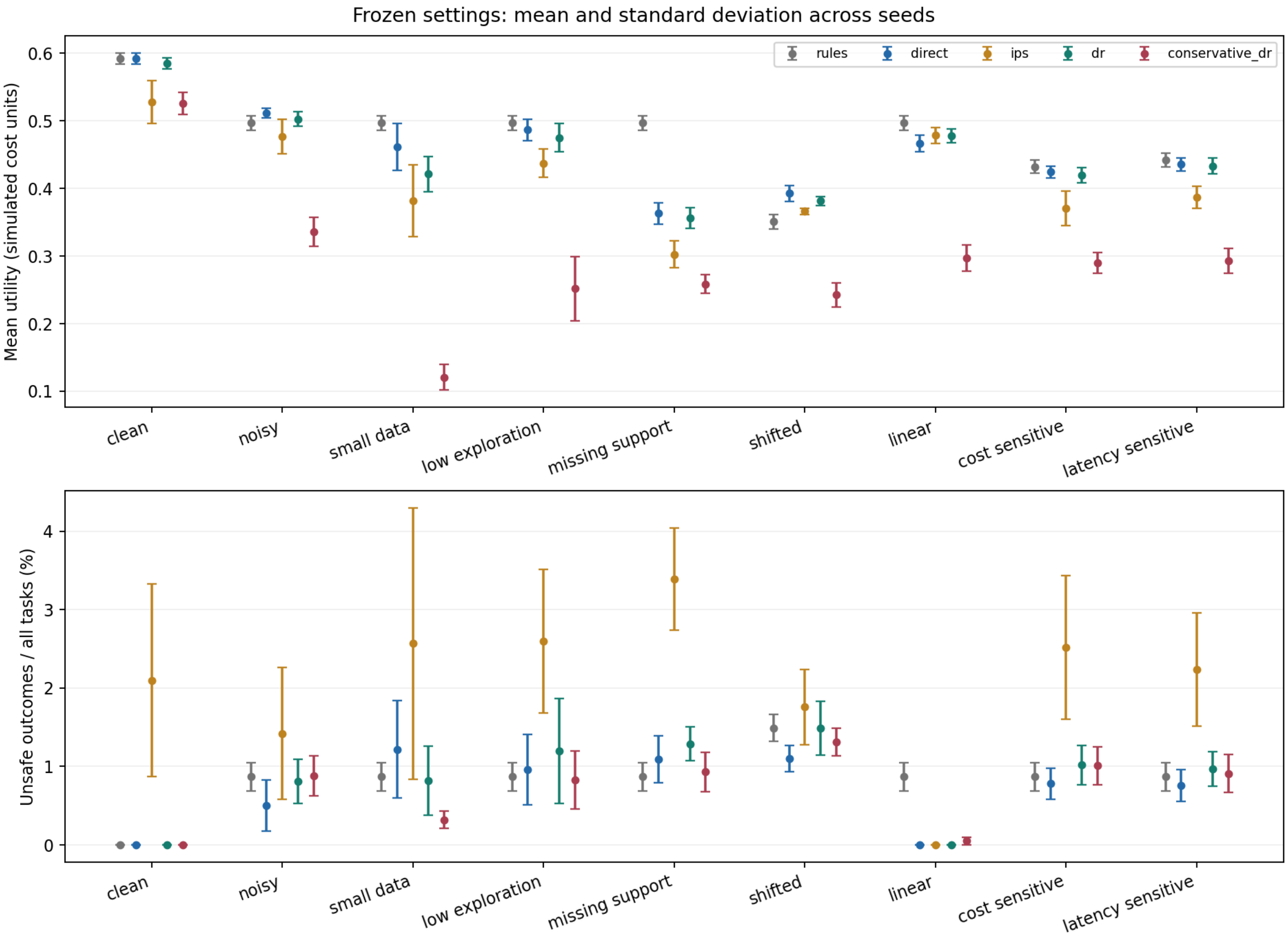

## 7.2 DR is more useful as a value estimator

Table 2 reports OPE mean absolute error against independent reset execution on the same test tasks. Cases count identified target-policy/seed pairs, not independent studies. The targets are the seven nontrivial policies listed in the experiment protocol.

| Setting | Direct | IPS | SNIPS | DR | Cases |
|---|---|---|---|---|---|
| clean | 0.0019 | 0.0413 | 0.0261 | 0.0010 | 35 |
| cost sensitive | 0.0129 | 0.0249 | 0.0200 | 0.0116 | 35 |
| latency sensitive | 0.0156 | 0.0269 | 0.0205 | 0.0120 | 35 |

| Setting | Direct | IPS | SNIPS | DR | Cases |
|---|---|---|---|---|---|
| linear | 0.1039 | 0.0404 | 0.0297 | 0.0193 | 35 |
| low exploration | 0.0517 | 0.0731 | 0.0406 | 0.0328 | 35 |
| missing support | 0.0119 | 0.0216 | 0.0192 | 0.0078 | 25 |
| noisy | 0.0143 | 0.0309 | 0.0239 | 0.0132 | 35 |
| shifted | 0.0772 | 0.0240 | 0.0210 | 0.0187 | 35 |
| small data | 0.0590 | 0.0221 | 0.0201 | 0.0148 | 35 |

Under shifted conditions, direct-model error is 0.0772, while DR error is 0.0187. For the linear model, errors are 0.1039 and 0.0193. Small-data errors are 0.0590 and 0.0148. Across this fixed matrix, DR has the lowest mean absolute error among the four compared estimators in every condition. This is an empirical observation for these targets and generators, not a universal ordering of estimators. The complete-return controls in Section 8 specifically reverse this ordering in one setting.

The shifted experiment evaluates target policies with held-out logs collected under the shifted test distribution. It does not show that old-distribution logs alone identify a new-distribution value without covariate-shift assumptions. The outcome model remains trained on the original noisy distribution, making residual correction useful while the current test logger supplies valid support.

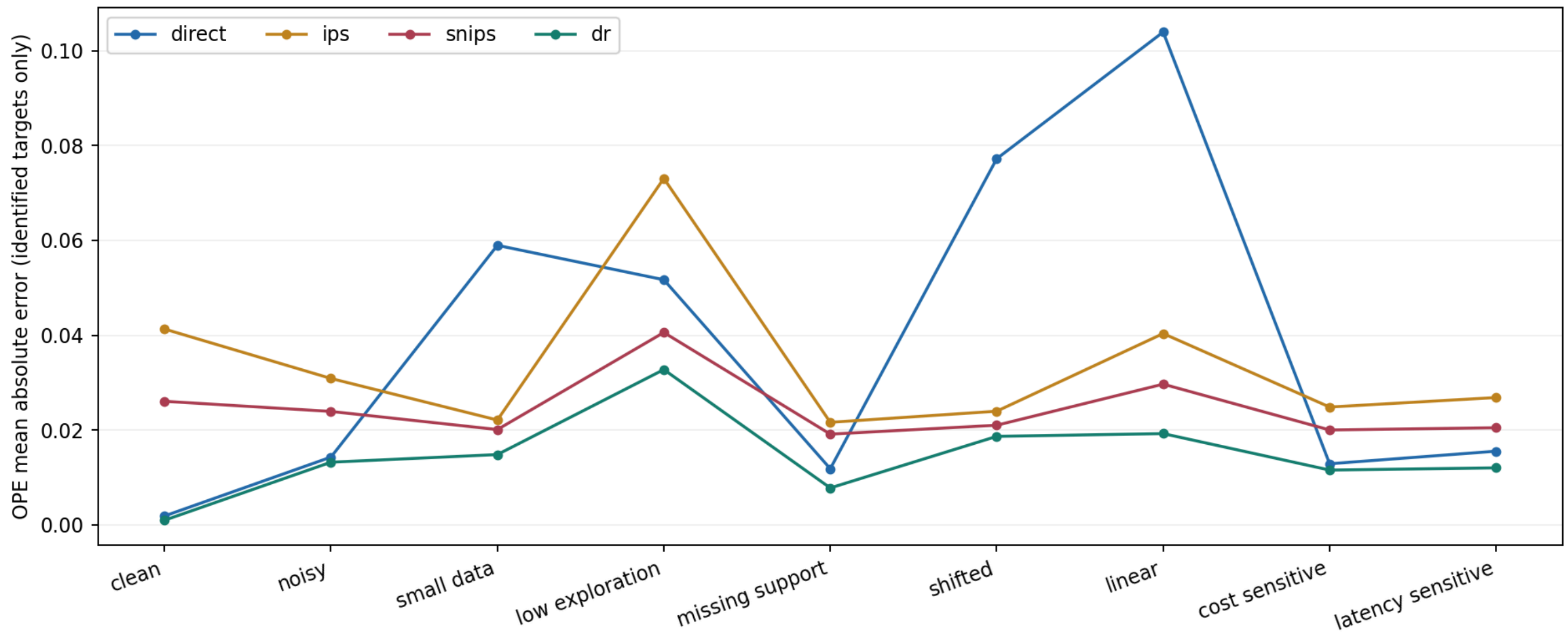

### 7.3 Abstention can lose utility without improving safety

Table 3 expands the noisy setting. Lower cost or risk must be interpreted together with coverage and success, rather than as a standalone improvement.

| Policy | Coverage % | Success/all % | Unsafe/all % | Cost/success | Utility |
|---|---|---|---|---|---|
| schema match | 81.31 | 55.93 | 0.00 | 0.1844 | 0.4562 |
| rules | 59.38 | 54.67 | 0.87 | 0.0592 | 0.4970 |
| direct | 73.07 | 57.77 | 0.50 | 0.0949 | 0.5118 |
| ips | 75.86 | 57.41 | 1.42 | 0.1127 | 0.4765 |
| dr | 74.08 | 57.86 | 0.81 | 0.1003 | 0.5027 |

| Policy | Coverage % | Success/all % | Unsafe/all % | Cost/success | Utility |
|---|---|---|---|---|---|
| conservative direct | 43.90 | 37.80 | 0.62 | 0.0653 | 0.3409 |
| conservative dr | 43.12 | 37.60 | 0.88 | 0.0592 | 0.3359 |
| oracle | 61.03 | 61.03 | 0.00 | 0.0510 | 0.5791 |

Conservative DR falls to utility 0.3359. Its evidence floor rejects many strata in a cross-product of domain, observed approval, scope set, resource count, service state, cache age, amount, and limit. Ensemble disagreement does not reliably detect stale approvals. The conservative variants consequently do not establish either a monotonic risk reduction or dominance at matched coverage. Their poor utility is a result, not an omitted failure.

The complete checked-tool baseline has zero unsafe business outcomes in this construction because its implementation validates preconditions. Learning can trade some of that conservative behavior for lower fees and higher coverage, but that trade is not an authorization relaxation. A practical deployment must specify a separate acceptable business-risk budget rather than assume that a weighted reward alone enforces it.

Figure 3 plots the held-out conservative-DR threshold sweep with cost shown by color and business risk shown separately. It is a set of sampled operating points, not a proven global Pareto frontier. Where a threshold produces no executions, conditional success is undefined; such seed points are excluded from that conditional curve and the accompanying JSON records their count. Thresholds were specified before test evaluation, not selected using this plot.

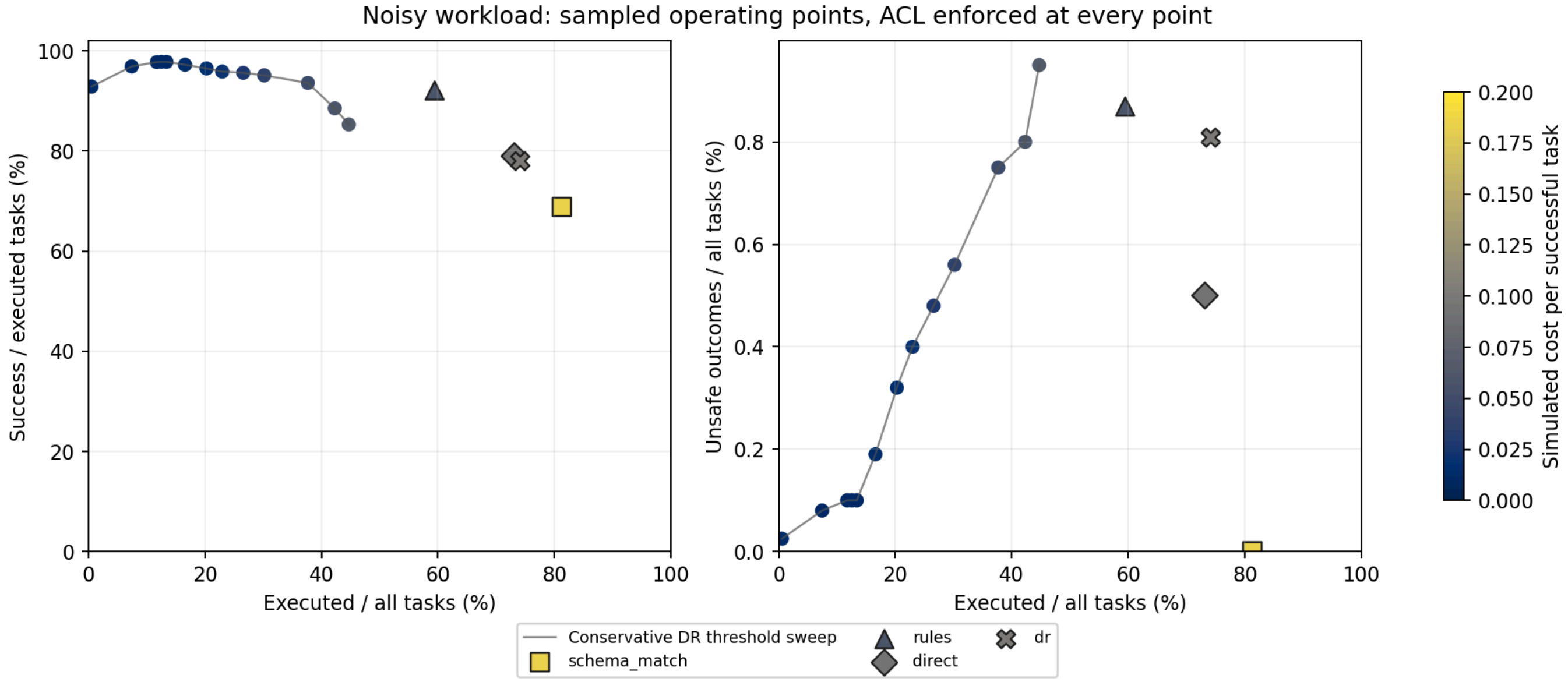

## 7.4 Missing support and protocol correctness

When three tools are never explored, full values for the rules and complete-tool policies are not identified for any of the five seeds. The target actions are legal; the problem is missing counterfactual evidence. Other restricted learned policies remain evaluable, but their reduced attainable value is visible in Table 1. We do not compare errors by inventing outcomes for the excluded tools.

A separate noisy MCP run uses 600 training, 200 calibration, and 400 test tasks. It performs 2542 actual tool calls through the stdio subprocess and negotiates protocol 2026-07-28. Measured round-trip median and 95th percentile are 0.755 and 0.975 milliseconds on the development machine. These are local transport timings, not production service latency or API cost. Replaying all 600 selected training calls through MCP produces zero semantic mismatches. Integration tests additionally compare local and MCP results for clean and shifted fixtures, including server-side revocation.

No unauthorized execution is observed in the evaluated policies. This is expected from the deterministic mediator and tested implementation; the policies share that same gate. It is not evidence that the learner itself understands authorization or that a real enterprise deployment has zero security risk.

# 8. Version-2 Controls and Public Evidence

## 8.1 A fair realized-return comparison

The first direct estimator predicts success and unsafe outcomes, then subtracts nominal fees and latency. A revoked call is free in the simulator, and a failed call can change actual latency or avoid resource exposure. The DR residual sees the realized total return. An apparent advantage can therefore reflect a weak cost model rather than a tool-specific counterfactual-learning effect.

We add two controls: a scalar regressor of complete realized return, and a multi-output regressor of realized success, unsafe outcome, fee, latency and extra-resource count, combined with the same utility weights. Full-return DR cross-fits the same scalar return model. All version-2 selectors use the same legal and empirically supported action boundary, the same feature interface, and the same model-family settings. A regression test where execution authority is revoked verifies that the full-return controls learn zero utility rather than subtracting a nominal fee.

Thirty new runs cover six settings and five fixed seeds, using 6,000 training and 2,000 test decisions each. OPE is compared on identical frozen target-policy cases within each setting. Table 4 reports mean absolute errors; these cannot be numerically conflated with Table 2, which uses the original target set.

| Setting | Nominal DM | Full DM | Component DM | Nominal DR | Full DR |
|---|---|---|---|---|---|
| clean | 0.0016 | 0.0016 | 0.0098 | 0.0010 | 0.0014 |
| noisy | 0.0179 | 0.0183 | 0.0155 | 0.0110 | 0.0118 |
| shifted | 0.0956 | 0.0948 | 0.0932 | 0.0262 | 0.0227 |
| linear | 0.0870 | 0.0139 | 0.0139 | 0.0240 | 0.0272 |
| cost sensitive | 0.0138 | 0.0178 | 0.0127 | 0.0092 | 0.0080 |
| latency sensitive | 0.0197 | 0.0166 | 0.0139 | 0.0114 | 0.0111 |

The linear setting changes the scientific conclusion. Full-return direct regression has MAE 0.0139, better than full-return DR at 0.0272. The first version's nominal-cost comparison had obscured this strong baseline. Under a shifted test environment, the ordering reverses: full-return direct and DR errors are 0.0948 and 0.0227. Thus the benefit of correction depends on the quality of the complete nuisance model; it is not an intrinsic superiority of DR. Full per-policy utility tables, component controls and seed dispersion are released.

The scalar-return control also changes the outcome parameterization: it is not the same function class as two separately clipped success/risk heads. This comparison adds a missing practical baseline; it does not isolate cost modeling as the sole causal source of the reversal. The realized-component control and released model specifications make that distinction inspectable.

## 8.2 Independent BFCL-derived tasks

We use the public BFCL live-multiple and live-irrelevance question files and official reference function names at pinned revision `61fc0608cfd831fcfbbaa676ebdfef0ed963eeda` [11]. Their dataset card declares Apache-2.0. These live queries originated outside our simulator. Of 1,935 rows, one question lacks an exactly matching reference ID and four contain duplicate function names. They are excluded transparently, leaving 1,930 tasks; one orphan reference is recorded rather than joined by row position. All source messages are preserved with explicit role labels, including the 86 source rows with system or assistant context. The context is task data, not executable host instructions. Original question text, reference arguments, and model weights are not redistributed.

We split connected components of normalized full function names or duplicate normalized queries. There is zero normalized function-name overlap between training, calibration and test. This is not a proof of zero semantic or service namespace overlap. Common functions connect a 1,544-row component, which is always assigned to training by a predeclared greater-than-half rule. Remaining components use seeded hash assignment. Consequently each of the five test splits has only 57--132 tasks and 20--31 groups. This severe reduction is reported rather than hidden by a random row split or favorable seed selection.

TF-IDF word and character features are fitted on training observations only. They measure query/schema similarity and schema complexity, not function IDs or answer labels. Reference answers and categories are absent from features; cross-fitting uses GroupKFold over training components. This task is exact function selection plus abstention, not BFCL's complete AST/argument/execution metric. No tool endpoint contained in the public text is executed.

The underlying labels are public full feedback. We deliberately reveal only one sampled action's outcome per learning decision with new epsilon-greedy probabilities, recorded exactly. These are semi-synthetic bandit logs, not historical BFCL or enterprise propensities. Native utility is +1 for a correct call, -1 for an incorrect call, and 0 for abstention; abstention accuracy is separately evaluated against the no-relevant-function labels. Native data have no injected fee or ACL. Two explicitly derivative settings inject an independent 20% function permission mask or a 25% zero-support mask, plus a schema-length fee proxy capped at 0.05. These transforms are not claims about real privileges or provider charges in BFCL.

Table 5 reports native results averaged across fixed splits. Balanced accuracy averages should-call and should-abstain class recall, preventing class imbalance from making an all-abstain policy look successful. Group-bootstrap intervals resample connected components, not individual questions, and are included in the released diagnostics rather than used to select thresholds.

| Policy | Selection accuracy % | Balanced accuracy % | Coverage % |
|---|---|---|---|
| always abstain | 69.22 | 50.00 | 0.00 |
| direct | 86.42 | 81.85 | 27.63 |
| dr | 84.09 | 79.83 | 28.83 |
| ips | 84.44 | 79.63 | 28.06 |
| tfidf | 70.47 | 69.23 | 43.00 |

The complete-return direct selector achieves 81.85% balanced accuracy versus 69.23% for TF-IDF and 79.83% for DR. This supplies independent public task evidence for a learned selector, but not evidence that DR is the best selector. Tool schemas and requests are real public benchmark material; business side effects remain untested in this arm.

A post-run dataset audit finds a structural shortcut: all 508 single-candidate rows in the selected BFCL categories are should-abstain tasks. Candidate count is an observable feature, but exploiting it is not language understanding. Without retraining, we therefore additionally report only held-out tasks with at least two candidates (Table 5b). Direct balanced accuracy falls to 76.80%, compared with 64.89% for TF-IDF. The result still contains signal beyond the one-candidate shortcut, but the reduction underscores that the full-set headline is not a general semantic-reasoning score. This subgroup diagnostic was added after the runs and is not misrepresented as a preregistered primary endpoint.

| Policy | Selection accuracy % | Balanced accuracy % | Coverage % |
|---|---|---|---|
| always abstain | 40.37 | 50.00 | 0.00 |
| direct | 74.19 | 76.80 | 54.04 |
| dr | 70.25 | 72.26 | 55.38 |
| ips | 71.47 | 74.17 | 53.12 |
| tfidf | 63.76 | 64.89 | 64.19 |

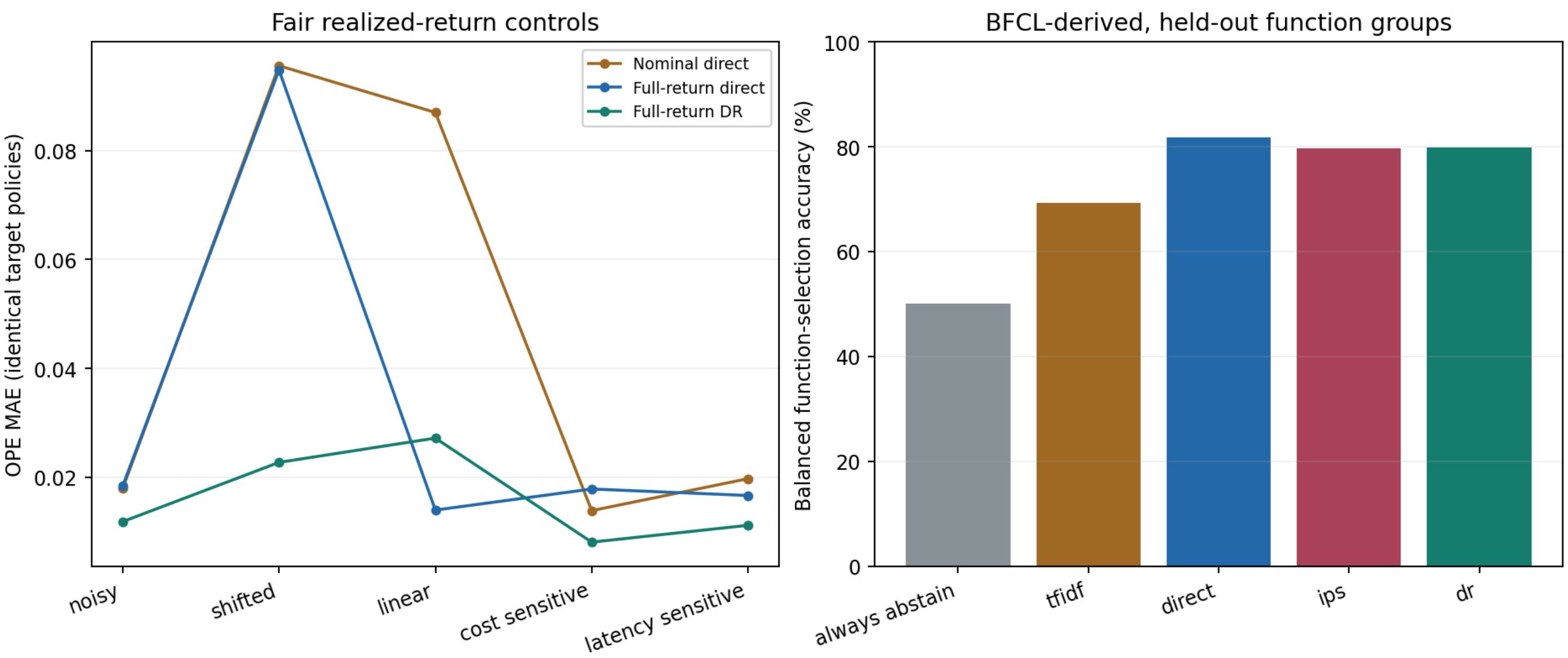

### 8.3 Actual local LLM baselines

We run pinned MLX 4-bit Qwen2.5-0.5B-Instruct and Qwen2.5-1.5B-Instruct models [12] on the same 200 SHA256-selected task IDs from the union of the five test splits. The union reuses tasks across splits; comparison to a learned selector is restricted to the seed for which that task is held out. Prompt design is fixed before test inference. Generation is greedy, with independent contexts, an 8,192 input-token cap and a 192 output-token cap. Invalid output and capped inputs are counted as errors, not silently excluded.

The models see only requests and schemas, and must emit one exact function name with JSON arguments or a null-name abstention. No reference labels are shown. We report exact selection and schema validity separately. Type aliases such as BFCL's `dict`, `int` and `any` are normalized for structural validation; schema validity is not evidence that values match the reference semantics. Weights, prompt hashes, token counts, runtime and task IDs are recorded.

| Model | Tasks | Should-call correct | Should-abstain correct | JSON valid % | Schema valid/known call % |
|---|---|---|---|---|---|
| qwen-0.5b | 200 | 46/72 | 0/128 | 95.00 | 92.47 |
| qwen-1.5b | 200 | 59/72 | 0/128 | 99.00 | 92.51 |

Neither model abstains on the 128 should-abstain questions; the remaining 72 tasks require a function call. Their overall selection accuracies are therefore much lower than their should-call recalls. This is a result for small quantized models under one fixed prompt, not a general statement about LLM tool-use capability. Unknown pretraining overlap with BFCL remains a contamination risk.

Table 7 uses exactly the same held-out task subsets for each learned method and LLM within a seed, then averages the five comparisons. The models generate parameters as well, whereas learned selectors choose names; only name selection is compared. It is not a frontier-model ranking or an official BFCL score.

| Matched model subset | Policy | Selection accuracy % | Balanced accuracy % |
| --- | --- | --- | --- |
| qwen-0.5b | always abstain | 66.83 | 50.00 |
| qwen-0.5b | direct | 85.95 | 82.36 |
| qwen-0.5b | dr | 83.76 | 80.44 |
| qwen-0.5b | llm | 23.19 | 34.33 |
| qwen-0.5b | tfidf | 70.90 | 70.41 |
| qwen-1.5b | always abstain | 66.83 | 50.00 |
| qwen-1.5b | direct | 85.95 | 82.36 |
| qwen-1.5b | dr | 83.76 | 80.44 |
| qwen-1.5b | llm | 27.68 | 41.73 |
| qwen-1.5b | tfidf | 70.90 | 70.41 |

## 8.4 Identifying policy changes rather than absolute values

Let $d(a \mid z) = \pi(a \mid z) - \pi_0(a \mid z)$, and let $U(z) = \{a : \mu(a \mid z) = 0\}$ be unsupported actions. Suppose outcomes lie in $[l, u]$, with no additional restrictions on unsupported conditional means. The estimand is the incremental value

$$\Delta = \mathbb{E}_z \sum_a d(a \mid z) q(z, a).$$

The following proposition specializes established deficient-support and baseline-restriction reasoning [3,4] to a linear contrast of two tool policies. Its contribution here is the explicit contrast-level condition and executable bounds, not a claim of priority over general partial-identification theory.

**Proposition (contrast support).** Under randomized logging and observation of the complete decision context, the contrast is point-identified for arbitrary bounded outcome models if and only if $d(a \mid z) = 0$ on $U(z)$ almost surely. Its sharp partial-identification width otherwise is

$$W = (u - l)\mathbb{E}_z \sum_{a \in U(z)} |d(a \mid z)|.$$

To see this, split the value sum into supported and unsupported terms. Supported conditional means are identified by the randomized logger. On unsupported terms, write $d_+ = \max(d, 0)$ and $d_- = \min(d, 0)$. Their lower and upper contributions are respectively $d_+ l + d_- u$ and $d_+ u + d_- l$. Assigning the endpoints independently to unobserved conditional means attains both bounds without changing the observed distribution. Subtracting yields $W$. When both policies select the same unsupported action, its coefficient is zero and it cancels exactly. If a nonzero unsupported coefficient occurs with positive probability, the two endpoint worlds establish non-identification. This proof does not claim priority over established partial-identification arguments; it exposes the relevant support condition for this tool-policy comparison.

Our implementation estimates only the supported contribution using a paired DR residual and retains the unsupported lower/upper terms explicitly:

$$\widehat{\Delta}_S = \frac{1}{n} \sum_i \left[ \sum_{a \notin U_i} d_i(a)\hat{q}_i(a) + \frac{d_i(a_i)}{\mu_i(a_i)}(r_i - \hat{q}_i(a_i)) \right].$$

The model and policies are frozen independently of the evaluation logs. For deterministic policies, disagreement fallback preserves the target where target and baseline agree or both selected actions have positive logging probability; otherwise it retains the baseline action. This makes the contrast supported without pretending that either absolute value is known. The baseline must itself obey the current ACL, but need not have certified business utility. Fallback preserves its risks as well as its behavior.

For bounded independent logging on fixed observed contexts, let $R_i$ be the range of possible importance-weighted residual corrections. Hoeffding's inequality gives a two-sided sampling radius

$$\epsilon_\delta = \sqrt{\frac{\log(2/\delta)}{2n^2} \sum_i R_i^2}.$$

The reported interval adds this radius to the estimated partial-identification endpoints. Shared-action rows have exactly zero range. The guarantee is conditional on the observed contexts, frozen policies and nuisance model, and independent randomized sampling. It is not a guarantee for a future distribution, human interactions or a data-dependent search over many policies.

In the five BFCL-derived support-gap runs, both separate absolute values remain unidentified for the comparisons in Table 8. Disagreement fallback identifies the relative change in all five, unlike blanket substitution of abstention. The zero width refers to structural identification, not zero sampling error.

| Comparison against TF-IDF | Both absolute values identified | Gain point-identified | Mean identification width |
| --- | --- | --- | --- |
| dr | 0/5 | 0/5 | 0.1315 |
| blanket support abstain | 0/5 | 0/5 | 0.2111 |
| disagreement fallback | 0/5 | 5/5 | 0.0000 |

Conservative calibration never yields a strictly positive lower bound, so no automatic upgrade is certified in any of the 15 public runs. This negative result is important: the method fixes the estimand's support requirement, but does not manufacture data or solve the sample-complexity problem. A deployable improvement still needs sufficient independent evidence and explicit risk budgets.

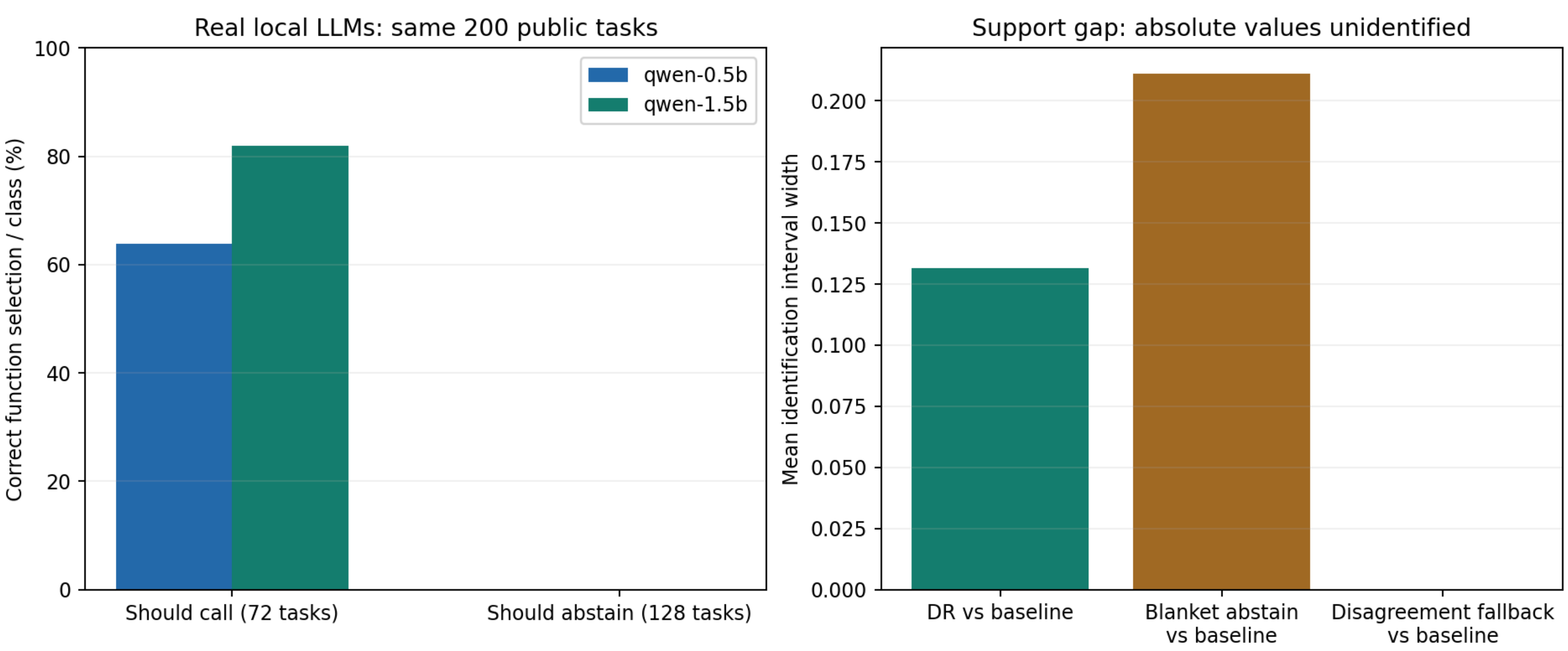

## 9. Discussion and Limitations

**Decision learning and policy evaluation are different deliverables.** A useful first deployment of this architecture may be better logging and policy evaluation around an existing direct ranker, not replacement of that ranker with DR pseudo-outcome regression. The experiment provides no reason to discard the stronger direct baseline simply because the paper's motivation mentions counterfactual learning.

**Sparse support is an engineering constraint, not an estimator detail.** Exploration takes place only among approved sandbox actions. Neither a new ACL grant nor a powerful outcome model creates evidence for a never-tried action. Real deployments need an explicit exploration or evidence-acquisition policy, and potentially a human approval path, before extending autonomous action sets.

**The conservative heuristic is inadequate as a safety mechanism.** Its evidence floor is too coarse to represent all business risks and too fragmented to retain useful coverage. A subsequent study should compare calibration under matched coverage, risk-constrained policy selection, and state-dependent information acquisition. Such changes should be specified before collecting the next test set, rather than tuned against the current matrix.

**External evidence is improved but still limited.** The executable environment has eleven tools and three synthetic domains. The new public BFCL-derived arm adds independent text, tool schemas and actual LLM inference, but evaluates selection rather than executed business goals. Function-group disjointness is not complete semantic novelty, and small test groups limit precision. Neither arm tests thousands of executable MCP tools, paid provider costs, interactive approval, a persistent LLM planner, or multi-step policy improvement. No production-agent or official BFCL, tau-bench or AgentAbstain score is claimed.

**Authority is simplified.** The access-control model supports principal/group grants, resource inheritance, deny precedence, and revocation snapshots. It does not implement a full enterprise IAM system, transitive group graphs, distributed revocation, or formal verification. The server manifest is trusted. A caller who can replace that manifest is outside the threat model.

**Utility design is subjective.** Simulated fees, unsafe-outcome penalties, exposure proxies, and a zero-utility abstain baseline encode particular operator preferences. Sensitivity settings expose some consequences but do not establish universal weights. Real deployments must measure costs and user utility, include the cost of asking or escalating, and distinguish recoverable from irreversible side effects.

**The original comparator limitation materially affects conclusions.** The nominal-cost baseline remains in the frozen first study for transparency. Complete-return and realized-component controls are now implemented and tested, and reverse the original OPE ordering in the linear setting. This does not prove that the new baseline is universally optimal; richer models and alternative regularization remain unexplored. Reporting the reversal is more informative than preserving a headline that depended on an incomplete control.

**Statistical conclusions are preliminary.** Five seeds and 200 bootstrap resamples support an inspectable first study, not a definitive method ranking. Hyperparameter search is limited, classifiers are not probability-calibrated on an independent action-level validation target, and interval coverage is not uniformly guaranteed after selection. Reused task templates and coupled setting seeds are explicit restrictions on generalization and statistical independence.

## 10. Reproducibility and Data Handling

The public repository contains the simulator, authorized candidate generation, exact logger, three learners, conservative variants, four OPE estimators, MCP server/client, tests, fixed experiment protocol, per-run summaries, and generated tables and figures. The full experiment can be recreated without an API key:

```
python3 -m venv .venv
.venv/bin/python -m pip install -r requirements.txt -r requirements-dev.txt
```

```
.venv/bin/python -m unittest discover -p 'test_*.py' -v
.venv/bin/python suite.py --out results/reproduction
```

The version-2 controls and public data are reproduced separately:

```
.venv/bin/python v2_experiments.py --out results/v2-reproduction
.venv/bin/python -m pip install -r requirements-llm.txt
.venv/bin/python public_llm.py --model 0.5b --out results/llm-small-reproduction
.venv/bin/python public_llm.py --model 1.5b --out results/llm-large-reproduction
.venv/bin/python verify_v2.py
```

MLX inference requires Apple silicon; recorded model outputs and all other experiments can be inspected on other platforms. The public dataset downloader uses pinned revisions and content hashes. License, attribution, exclusions and split assignments are documented in the repository. Original BFCL prompts, reference arguments, raw generated text and model weights remain in local caches; public outputs contain task IDs, scores, hashes and aggregate diagnostics.

A pre-publication input audit found that the first version-2 adapter had kept only user messages and omitted other public roles. Those preliminary results were not published as the official extension. We preserved them locally and reran the complete version-2 matrix and both models after restoring the full role-labeled context, with unchanged seeds, model weights, prompt template, thresholds and token caps. Source hashes identify the corrected adapter used by both the learned selectors and the local models.

For the protocol check:

```
.venv/bin/python run.py run --backend mcp --scenario noisy \
  --train-size 600 --calibration-size 200 --test-size 400 \
  --bootstrap 100 --seed 7 --out results/mcp-reproduction
.venv/bin/python run.py replay --backend mcp \
  --log results/mcp-reproduction/train.jsonl
```

Raw generated logs remain local by default. Published summaries remove local output paths and expose only synthetic metadata and aggregate outcomes. Raw-log hashes are recorded for provenance, but measured runtimes make byte-for-byte hashes machine-specific; semantic replay ignores that one measurement. The working paper's tables are generated from committed JSON summaries instead of being independently transcribed. All figures are generated from the same data.

## 11. Conclusion

Counterfactual tool selection is useful only when its execution, logging, and evaluation contracts are explicit. Our implementation keeps deterministic authorization separate from learned value and separates legal availability from historical support. Full realized-return controls overturn a favorable DR comparison in one setting, while correction remains helpful under others. Independent BFCL-derived tasks and real local LLMs add external evidence without turning a selection benchmark into a production claim. At the estimand level, unsupported common actions need not prevent identifying a policy change; a disagreement-preserving fallback makes this condition operational. Its bounds remain too wide to certify the changes in our public-data study. The resulting lesson is conditional, testable and narrower than claiming a universally better ranker: model the full return, measure both action and abstention performance, and distinguish identifiable changes from unsupported absolute predictions.

## Generative-AI Assistance

GitHub Copilot assisted with research design, software implementation, experiment execution, analysis, and manuscript drafting. Tables and numeric claims were generated from executable experiments and checked against the published artifacts; they were not generated as hypothetical results. Automated tests and artifact checks do not replace independent human scientific review. Responsibility for the accuracy, originality, references, and interpretation of the submitted work remains with the named human author. The initial synthetic study uses no generative model as an agent; version 2 separately evaluates the two explicitly identified Qwen2.5 models as experimental baselines.